\documentclass[sigconf,nonacm]{acmart}

\usepackage{amsmath}
\usepackage{graphicx}
\usepackage{tabularx}
\usepackage{booktabs}
\usepackage{multirow}
\usepackage{siunitx}
\usepackage{subcaption}
\usepackage{paralist}
\usepackage{caption}
\usepackage{footmisc}
\usepackage[table]{xcolor}
\definecolor{cellgreen}{RGB}{158, 210, 150}   
\definecolor{cellred}{RGB}{235, 150, 140}      

\newboolean{anonymous}
\setboolean{anonymous}{false} 

\hypersetup{
  colorlinks=true,
  linkcolor=blue,
  citecolor=blue,
  urlcolor=blue
}

\newcommand{\figref}[1]{\hyperref[#1]{Fig.~\ref*{#1}}}
\newcommand{\secref}[1]{\hyperref[#1]{Sec.~\ref*{#1}}}
\newcommand{\tabref}[1]{\hyperref[#1]{Table~\ref*{#1}}}
\renewcommand{\eqref}[1]{\hyperref[#1]{Eq.~(\ref*{#1})}}

\begin{document}

\title{SPECS: Speciated Evolutionary Circuit Synthesis}

\ifthenelse{\boolean{anonymous}}{
  \author{Anonymous Authors}
  \affiliation{%
    \institution{Anonymous Institution(s)}
    \country{}}
  \renewcommand{\shortauthors}{Anonymous}
  \setcopyright{acmlicensed}
  \copyrightyear{2026}
  \acmYear{2026}
  \acmDOI{XXXXXXX.XXXXXXX}
  \acmConference[ICCAD '26]{International Conference on Computer-Aided Design 2026}{November 8-12, 2026}{San Jose, CA}
  \widowpenalty=100     
  \clubpenalty=100      
}{
    \author{Yağız Gençer\textsuperscript{1, 2}, Stefan Uhlich\textsuperscript{1}, Andrea Bonetti\textsuperscript{1},\linebreak Arun Venkitaraman\textsuperscript{1}, Chia-Yu Hsieh\textsuperscript{1}, Lorenzo Servadei\textsuperscript{1,3}}
    \affiliation{%
    \institution{\textsuperscript{1}\textit{Sony Group Corporation, Switzerland} \quad\quad\quad
    \textsuperscript{2}\textit{EPFL, Switzerland} \quad\quad\quad
    \textsuperscript{3}\textit{TU Munich, Germany}}}
    \renewcommand{\shortauthors}{Gençer, Uhlich, Bonetti, Venkitaraman, Hsieh, Servadei}
}

\begin{abstract}
We propose SPECS, a genetic algorithm for automated analog circuit synthesis with joint topology and sizing optimization. SPECS is inspired by NeuroEvolution of Augmenting Topologies (NEAT)~\cite{NEAT}, an evolutionary algorithm originally developed to synthesize neural networks. By reformulating the genome representation and adapting the genetic operators to the analog circuit domain, we successfully transfer the core principles of NEAT to analog circuit synthesis. Circuit-specific wiring constraints are incorporated to ensure valid and physically meaningful designs throughout the evolutionary process, and speciation is used to preserve innovation while maintaining population diversity. We evaluate the proposed method on a set of computational circuit synthesis tasks consisting of square, cube, square root, and cube root functions. Experimental results demonstrate that SPECS outperforms benchmark methods across all tasks in both solution quality and reliability. The synthesized circuits and their schematics are available in the supplementary repository\footnote{\label{fn_repo}\url{https://anonymous.4open.science/r/specs-artifact-637A/}}.
\end{abstract}


\maketitle

\section{Introduction}
\label{sec:introduction}
Despite decades of research, analog circuit design remains a difficult and time-consuming process that still depends heavily on expert knowledge and manual tuning \cite{jespers2017systematic,razavi2017analogcmos}. Unlike digital design, where abstraction and automation have enabled highly scalable synthesis and optimization flows \cite{wang2009electronic,lavagno2018eda}, analog design is characterized by complex, non-convex search spaces, strong coupling between design variables, and strict physical constraints that have limited the adoption of fully automated design methodologies.

Evolutionary algorithms have long been explored as a promising approach for analog circuit synthesis, since they can optimize simulation-based, non-differentiable objectives while handling both topology choices and continuous component parameters \cite{analog_evo, app_evo}. Nevertheless, realizing this potential in an effective and scalable way remains difficult. Many existing methods either assume fixed circuit topologies \cite{sizing_1, sizing_2} or restrict the search to combinations of predefined building blocks \cite{building_block}. In more flexible search settings, efficiently exploring the solution space while avoiding premature convergence remains a key challenge.

In this paper, we introduce SPECS, a novel genetic algorithm for automated analog circuit synthesis inspired by NeuroEvolution of Augmenting Topologies (NEAT)~\cite{NEAT}. We adapt NEAT's core principles, such as incremental topology growth and speciation, to the analog circuit domain by redefining the genome representation, node and connection semantics, and genetic operators to act on circuit elements and interconnections. Starting from minimal structures, SPECS simultaneously evolves topology and component parameters. Speciation maintains population diversity and shields novel circuit motifs from premature elimination by more mature topologies, leading to more consistent discovery of high-performing solutions across independent runs.

The main contributions of this paper are:
\begin{itemize}
    \item A novel genetic algorithm for analog circuit synthesis that jointly evolves circuit topology and component parameters,
    \item A domain-specific genome representation and genetic operators tailored to analog circuits, incorporating wiring constraints to ensure validity throughout evolution,
    \item A speciation-based search strategy that protects novel circuit candidates and grows complexity only when beneficial,
    \item An experimental evaluation on computational circuit synthesis tasks consisting of squaring, cubing, square root, and cube root functions, demonstrating improved solution quality and reliability compared to benchmark methods.
\end{itemize}
The remainder of this paper is organized as follows. \secref{sec:related_work} reviews related work. \secref{sec:methodology} describes the proposed algorithm in detail. \secref{sec:experiments}  presents experimental results and comparisons with benchmark methods. Finally, \secref{sec:conclusion} summarizes the findings and discusses directions for future work.

\section{Related Work}
\label{sec:related_work}
Analog circuit synthesis generally involves two main tasks: (i) determining the circuit topology and (ii) selecting component parameter values. Substantial progress has been made on component sizing when the topology is fixed in advance \cite{noren_ross_ga,barari2014,yengui2012_ga_sqp,rashid2023_eval_ea,rashid2024_ml_global_opt,kwon2023_cga}, but identifying the correct topology remains a challenging and largely open problem.

Several evolutionary approaches have been proposed for topology synthesis. Early methods evolved both topology and component parameters using genetic algorithms, often incorporating constrained building blocks or subcircuit-aware crossover techniques to enhance the validity of generated designs \cite{kruiskamp1995_darwin,das2007_gapsys}. Genetic programming has also been applied, demonstrating that topology and sizing can be discovered simultaneously through evolutionary search \cite{koza1996_analog_gp, genetic_programming}. Meanwhile, grammar-based strategies decode variable-length chromosomes into circuit netlists, with recent extensions introducing modularity and homology to reduce crossover disruption and improve substructure reuse \cite{acid_ge,acid_mge}. Graph-based autoregressive formulations have been explored, generating circuits incrementally as graphs before translating them into SPICE netlists~\cite{graco_es}. Netlist-level evolution emerged as an alternative, with genetic operators applied directly to normalized SPICE netlists, eliminating the need for an explicit genome encoding \cite{spicemixer}.

SPECS differs from these approaches through its circuit-native genome representation and mutation operators that explicitly account for wiring constraints, enabling joint topology and sizing search within a single framework while preserving electrically valid structures. SPECS also promotes exploration through speciation, which adaptively groups genomes by genetic similarity to protect novel candidates. Related ideas appear in island-based parallel evolutionary methods \cite{pga_1,pga_2,pga_3}, where fixed subpopulations evolve independently and exchange individuals via migration. In contrast, SPECS reassigns genomes to species at each generation, allowing niches to emerge organically from the evolving population rather than being fixed in advance. This allows species to grow, shrink, or disappear over time, preventing less mature innovations from being exposed to premature competition outside their niche.

\section{Methodology}
\label{sec:methodology}
We now describe SPECS, the proposed method for automated analog circuit synthesis. First, the genome representation used to encode analog circuits is introduced. Next, the genetic operators, namely crossover and several mutation operators, are detailed, explaining how they act on a genome while enforcing the wiring constraints required for valid analog circuits. Finally, an overview of the complete algorithm is presented, demonstrating how a population of circuits evolves to solve the analog circuit synthesis task.

\subsection{Genome Representation}
In the context of genetic algorithms, a genome encodes a candidate solution to an optimization task in a structured representation that enables the application of genetic operators such as crossover and mutation. A genome consists of a collection of genes, each representing a specific aspect of the solution.

In NEAT, which evolves neural networks, genomes are composed of node and connection genes. A node gene stores a unique identifier and its type (input, output, or hidden). A connection gene represents a directed edge between two nodes and stores the input and output node IDs, connection weight, and a Boolean flag indicating whether it is enabled. Each connection is also assigned an innovation ID, which uniquely identifies that structural change: if a connection between a given pair of node genes appears for the first time anywhere in the population during a run, a new innovation ID is created; otherwise, the existing ID is reused.

This genome structure is not directly applicable to analog circuits due to fundamental topological differences. In circuits, the equivalent of a node is a component, but connections are not defined directly between components. Instead, they are established between nets and component pins, with each component having multiple pins. Moreover, circuit connections are undirected and weightless; tunable parameters are associated with the components themselves, such as resistance values or transistor dimensions.

\begin{figure}[t]
    \centering
    \includegraphics[width=\linewidth,trim=0 25 0 0]{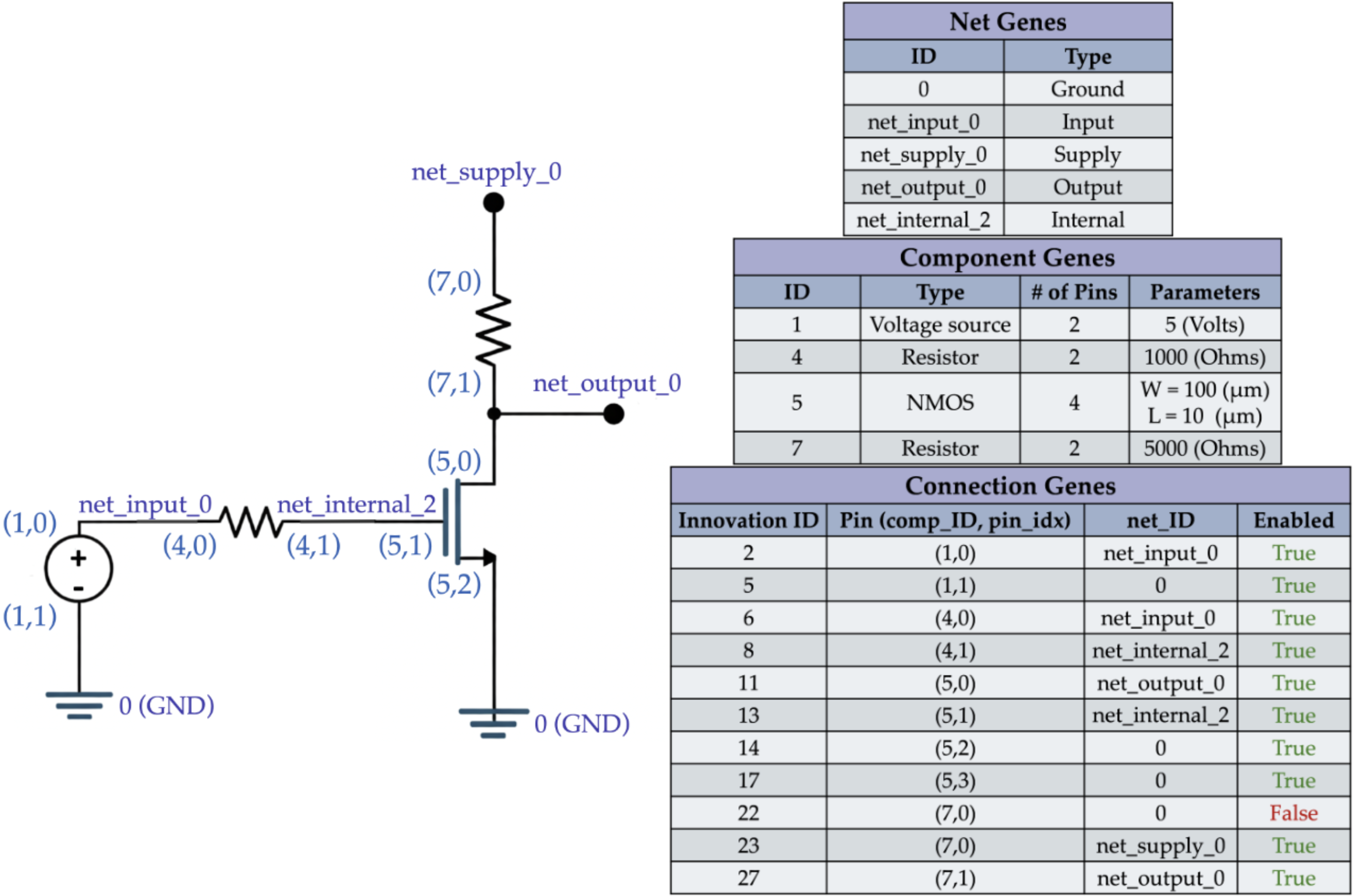}
    \caption{Genome representation of an example circuit.}
    \label{fig:genome_rep}
    \vspace{-0.4cm}
\end{figure}

To account for these properties, we propose a genome representation consisting of three gene types: component genes, net genes, and connection genes, as shown in \figref{fig:genome_rep} for an example circuit. A component gene stores the component ID, type (e.g., MOS, BJT, resistor), number of pins, and component parameters. A net gene stores a net ID and type (input, output, supply, ground, or internal). A connection gene stores an innovation ID, a component-pin tuple of the form \emph{(component ID, pin index)} that identifies which pin of which component is being connected, the net ID to which the pin is connected, and a Boolean \emph{Enabled} flag.

Component, internal net, and innovation IDs are assigned globally throughout evolution. Whenever a new component is created, it receives the next available component ID, and likewise for new internal nets. For connection genes, the identifying pattern is the pair of the component-pin tuple and the net ID. If a connection between a given component pin and a given net appears for the first time during a run, a new innovation ID is assigned; otherwise, the previously assigned innovation ID is reused. Consequently, identifiers within a genome need not be consecutive, since some IDs may correspond to structures created elsewhere in the population or earlier in evolution. When a connection gene becomes invalid due to certain genetic operators, it is disabled instead of removed, preserving evolutionary history. When aligned according to their innovation IDs, the matching connection genes of two genomes provides a meaningful measure of genome similarity, which is used for speciation (see \secref{sec:proposed_algorithm}).

\figref{fig:genome_rep} illustrates an example circuit and its genome. The pin at index $i$ of component $x$ is annotated as $(x,i)$, and the ground net is always assigned net ID 0. Note that the bulk pin of the NMOS transistor (pin index 3) is not shown in the schematic; nevertheless, it is represented in the genome by a connection gene linking pin $(5,3)$ to the ground net with innovation ID 17. In addition, the disabled connection gene with innovation ID 22 indicates that pin index 0 of the resistor with component ID 7 was previously connected to the ground net. This connection has since been disabled, and the current active connection of the same pin to the supply net is represented by a new connection gene with innovation ID 23.

\subsection{Genetic Operators}

SPECS employs three types of genetic operators: \begin{inparaenum}[(i)]
\item crossover,
\item structural mutations, and
\item sizing mutations.
\end{inparaenum}
Structural mutations modify circuit topology, while sizing mutations adjust component parameters without altering connectivity. Unlike NEAT, where structural mutations primarily increase complexity, structural mutations in SPECS consist of constructive (\emph{Add Component} and \emph{Split Net}), simplifying (\emph{Delete Component}), and rearranging (\emph{Rewire}) operators. This is required, as the performance of analog circuits is highly topology-dependent: connections do not carry tunable weights, and component parameters alone cannot compensate for structurally unfavorable designs. To address this, the operator set of SPECS allows not only controlled topology growth but also revision of previous design decisions.

\subsubsection{Crossover}
Given two parent genomes, crossover produces a single offspring. The topology of the fitter parent, in terms of the task-specific fitness defined in \secref{sec:experiments}, is inherited directly, reflecting the observation that structural quality is strongly correlated with fitness in analog circuit synthesis. For components shared by both parents (i.e., with the same component ID), the parameters are inherited from either parent with equal probability. Components present only in the fitter parent pass their parameters unchanged to the offspring. This mechanism preserves structural integrity while enabling parameter-level recombination between parents.

\subsubsection{Structural Mutations}
Structural mutations modify circuit topology via four operators: \emph{Add Component}, \emph{Delete Component}, \emph{Split Net}, and \emph{Rewire}.

A central requirement for structural mutations is to avoid creating floating nets. An internal net is floating if fewer than two component pins are attached to it. Input, supply and ground nets are floating if no pin is attached to them. Since the experimental test fixture connects the output net to ground through a load resistor (see \secref{sec:experiments}), an output net is also considered floating only if no pin is attached to it. We call a pin selection \emph{safe} if applying the corresponding operation does not create any floating nets. Thus, whenever pins are sampled in the steps below, they are selected at random from the set of safe choices when possible. If no safe choice exists, they are sampled from the set of all existing pins.

After each structural mutation, cleanup rules are applied if necessary: nets with no remaining pins are removed, and internal nets with a single remaining pin are removed after rewiring that pin to another existing net. This ensures that all resulting circuits remain electrically valid throughout evolution.

\paragraph{Add Component.}
A component type is sampled at random from the predefined library and its parameters are randomly initialized within valid ranges. A new component ID is assigned and the component is inserted according to one of three randomly sampled modes, which differ in how many existing pins in the genome are detached and reconnected through the new component:
\begin{itemize}
    \item \textbf{Zero-detachment mode:} No existing pin is detached.
    
    \item \textbf{One-detachment mode:} One existing pin and one pin of the new component are selected. The existing pin is detached from its current net, and the two pins are connected through a newly created internal net.
    
    \item \textbf{Two-detachment mode:} Two existing pins and two pins of the new component are selected. Each detached existing pin is paired with one selected pin of the new component, and each pair is connected through a newly created internal net, creating two new internal nets.
\end{itemize}

After this initial insertion step, any remaining pins of the new component are connected to randomly sampled nets present in the genome. As a result, the operator introduces a new component while minimally disrupting the genome’s existing topology.

\paragraph{Delete Component.}
A randomly selected component gene and all its associated connection genes are removed from the genome. This enables topology simplification by removing components that no longer contribute to performance, allowing the search to recover from previously suboptimal design decisions.

\paragraph{Split Net.}
A splittable net is selected at random. A net is considered splittable if moving at least two of its pins to a new internal net does not leave it floating, which requires it to have at least four attached pins if it is internal, and at least three if it is an input, output, supply, or ground net. A new internal net is created, and a random subset of at least two pins of the selected net is reassigned to the new net. This partitions a highly connected electrical node by introducing a new net, enabling alternative current paths and more refined connectivity.

\paragraph{Rewire.}
A pin is selected at random, detached from its current net, and reconnected to a randomly selected existing net. This operator enables connectivity refinement without changing circuit complexity.

In this framework, detaching a pin means invalidating its connection gene. When a connection gene is invalidated due to a constructive operator, it is disabled (its \emph{Enabled} flag set to False), thereby preserving its evolutionary history. In contrast, simplifying and rearranging operators remove the gene directly, since they undo prior decisions and frequently revisit similar complexity levels, where accumulation of inactive genes would inflate genome size and distort structural similarity metrics.

Beyond the floating-net requirement, SPECS can be easily extended with domain-specific wiring constraints by restricting the set of allowed pins or nets during sampling. For example, to prevent the gate terminal of MOS transistors from being connected to supply or ground, nets of these types can be excluded from the candidate set whenever a gate terminal, identified by pin index, is to be connected. Such constraints require no modifications to the overall algorithm, making SPECS easily adaptable to problem settings with specific electrical requirements.

\subsubsection{Sizing Mutations}
Two sizing operators adjust component parameters while preserving topology.

\paragraph{Resample Parameters.}
A component is selected at random and all of its parameters are resampled independently and uniformly within their valid ranges. This mutation enables large parameter-space exploration and allows escape from poor local configurations.

\paragraph{Perturb Parameters.}
A component is selected at random and each of its parameters is perturbed by resampling uniformly within an interval of width 10\% of the full valid range, centered at the current value and clipped to allowable bounds. This mutation enables fine-grained local search around promising parameter configurations.

\begin{figure*}[!t]
    \centering

    \begin{subfigure}[t]{0.49\textwidth}
        \includegraphics[width=\linewidth]{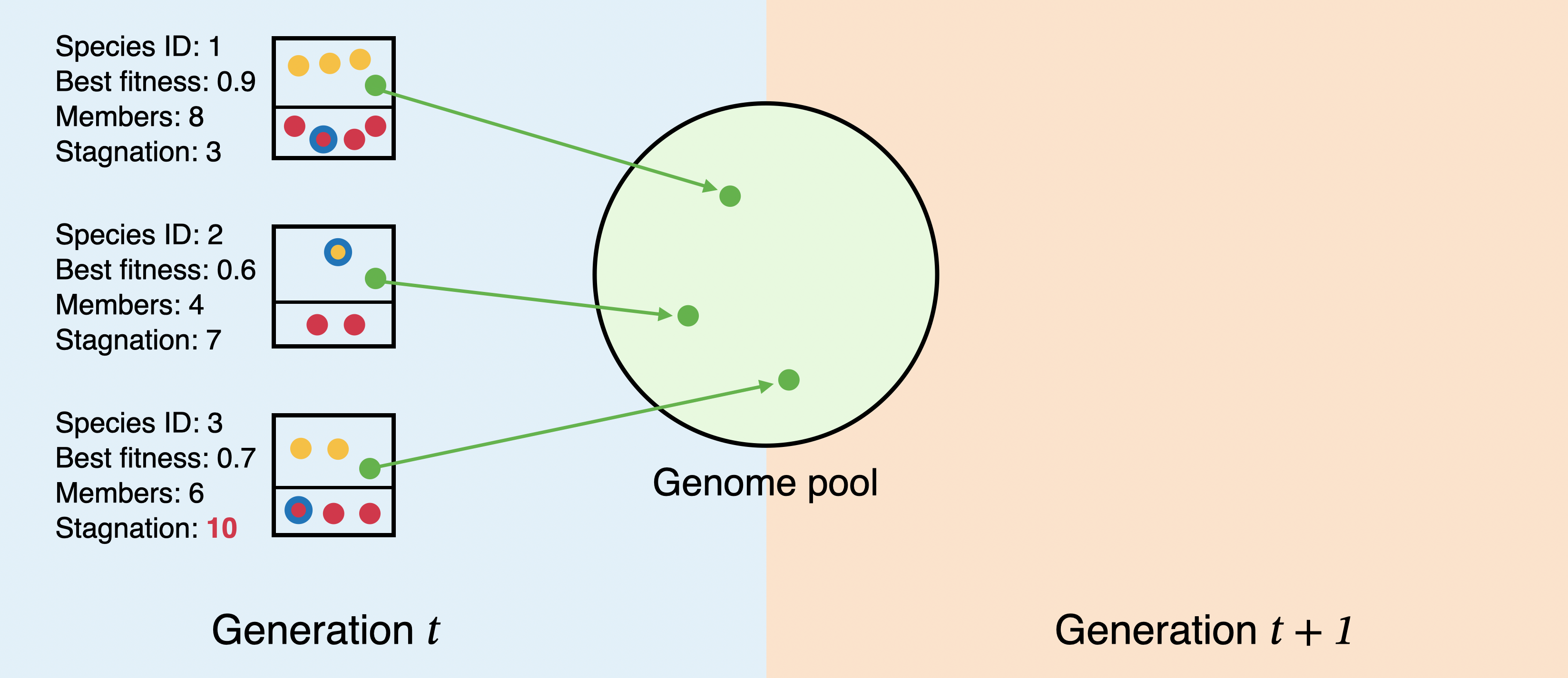}
        \vspace{-1.5em}
        \caption{}\label{fig:a}
    \end{subfigure}
    \hspace{0.01\textwidth}
    \begin{subfigure}[t]{0.49\textwidth}
        \includegraphics[width=\linewidth]{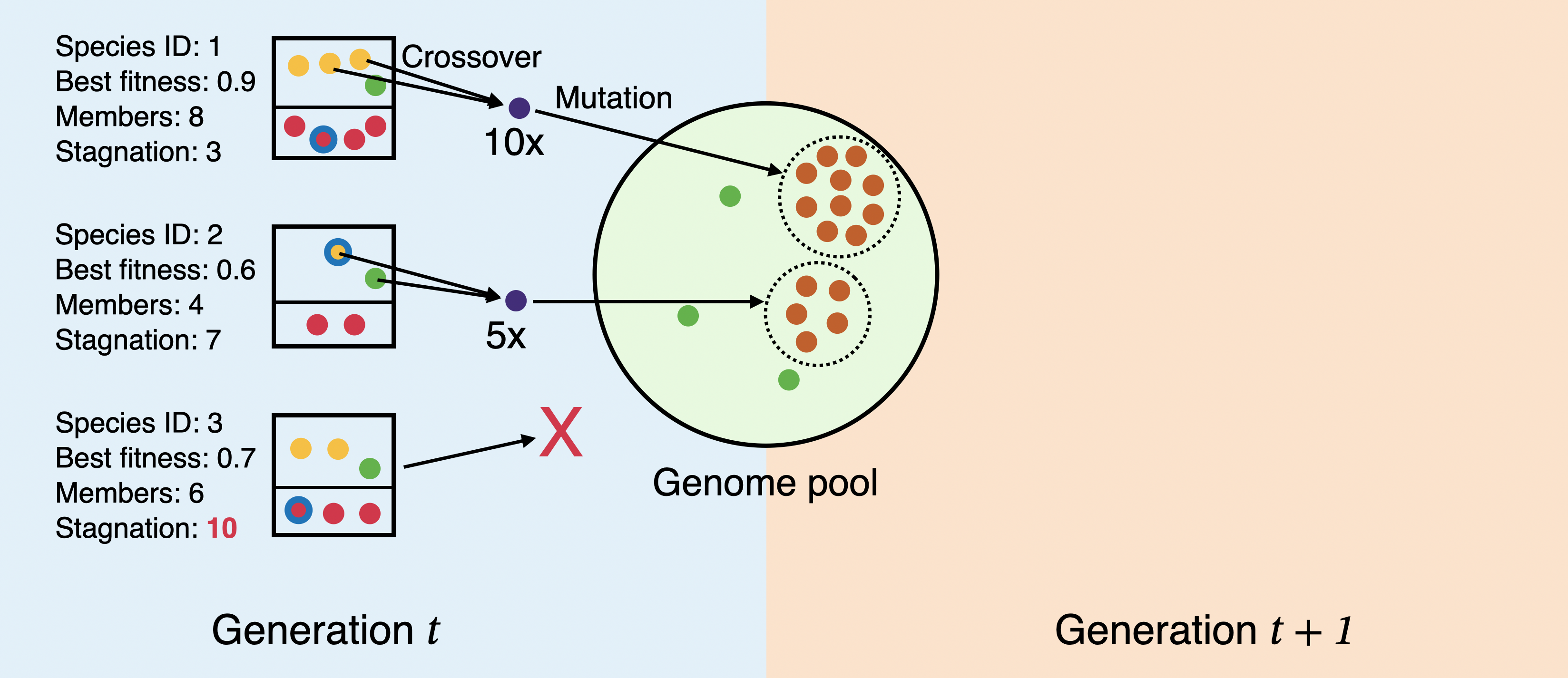}
        \vspace{-1.5em}
        \caption{}\label{fig:b}
    \end{subfigure}

    \vspace{0.4em}
    
    \begin{subfigure}[t]{0.49\textwidth}
        \includegraphics[width=\linewidth]{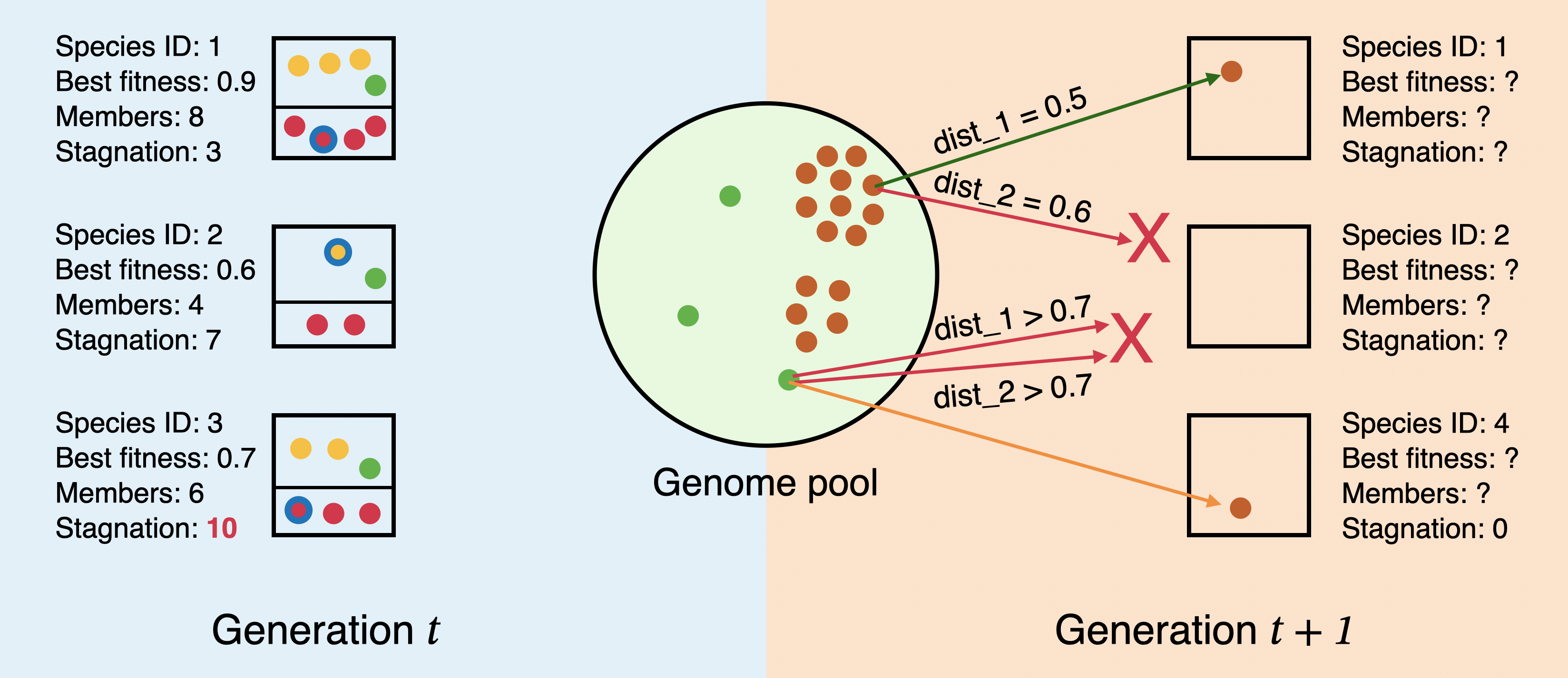}
        \vspace{-1.5em}
        \caption{}\label{fig:c}
    \end{subfigure}
    \hspace{0.01\textwidth}
    \begin{subfigure}[t]{0.49\textwidth}
        \includegraphics[width=\linewidth]{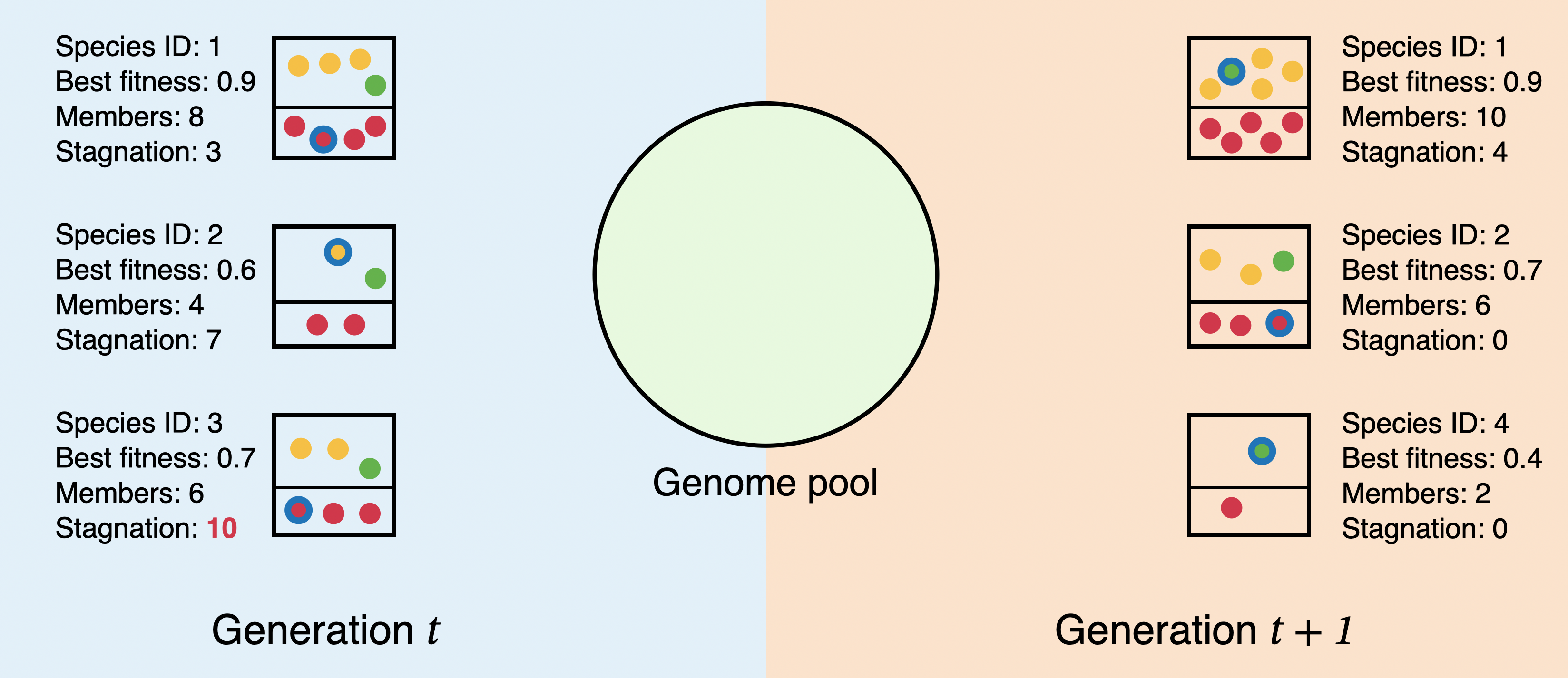}
        \vspace{-1.5em}
        \caption{}\label{fig:d}
    \end{subfigure}
    \vspace{-0.3cm}
    \caption{Reproduction mechanism in SPECS.
    (a) Elite (green) genomes are copied unchanged from each species to the genome pool.
    (b) Non-stagnated species generate offspring via crossover and mutation and contribute them to the genome pool. Parent selection is restricted to the top fraction (yellow and green), excluding the lower fraction (red) genomes.
    (c) Each offspring in the genome pool is assigned to the closest species representative (blue border) if the distance is within the compatibility threshold; otherwise it forms a new species.
    (d) Species statistics and representatives are updated for generation $t + 1$ after fitness evaluations. -- Parameters: population size = 18, elite survival fraction = 0.2, parent selection fraction = 0.5, compatibility threshold = 0.7, stagnation threshold = 10, elite size = 1.}
    \label{fig:algorithm}
\end{figure*}

\subsection{Distance between Two Genomes}

Let $g_1$ and $g_2$ denote two genomes. We define their distance as
\[
\delta(g_1,g_2) :=
\frac{c_1 E + c_2 D}{L}
+ c_3 \, \Delta_{\text{param}}
+ c_4 \, \Delta_{\text{comp}},
\]
where $E$ and $D$ are the numbers of excess and disjoint connection genes, as defined in NEAT via historical innovation IDs, and $L := \max(n_1, n_2)$ with $n_i$ the number of connection genes in genome $g_i$ normalizes for differences in genome size. Specifically, disjoint genes have innovation IDs present in one genome but not the other within the overlapping innovation ID range, while excess genes have innovation IDs beyond the maximum innovation ID of the other genome. The term $\Delta_{\text{param}}$ measures the average normalized parameter difference between matching components (those with identical IDs), with continuous parameters normalized by their valid ranges and discrete parameters contributing a unit penalty when mismatched. The term $\Delta_{\text{comp}}$ captures component-level topological differences as the number of components present in one genome but not the other, normalized by the total number of distinct components across both genomes. The coefficients $c_1, c_2, c_3, c_4$ control the relative importance of each term. This distance metric plays a key role in the speciation process, as described in \secref{sec:proposed_algorithm}.

\subsection{Proposed Algorithm}
\label{sec:proposed_algorithm}

We now describe the complete evolutionary procedure used by SPECS. The algorithm follows a generational evolutionary framework with speciation, elitism, crossover, and mutation operators aimed at topology selection and sizing. A schematic illustration of a transition from one generation to the next is provided in \figref{fig:algorithm}.

\paragraph{Initialization.}
Given a population size $N$, the initial generation is created by sampling $N$ genomes independently. Each genome contains a single randomly sampled component whose type and parameters are drawn uniformly from their valid ranges, with all pins connected to randomly sampled external nets (input, output, supply, or ground) and no internal nets.

As genomes are created, they are assigned to species. The first genome forms the first species and becomes its representative. For each subsequent genome, its distance to all existing species representatives is computed. If the minimum distance is below the compatibility threshold $\delta_t$, the genome joins the closest species; otherwise, it forms a new species and becomes its representative. After all genomes are assigned, generation 0 is complete.

\paragraph{Species Statistics and Stagnation.}
At each generation $t$, every species maintains:
(i) its best fitness value,
(ii) its number of members, and
(iii) a stagnation counter.

The stagnation counter tracks how many consecutive generations have passed without improvement in the species’ best fitness. If this counter exceeds a predefined \emph{stagnation threshold}, the species is considered stagnated and is prevented from generating new offspring in subsequent generations (e.g., Species 3 in \figref{fig:algorithm}(b), where stagnation $= 10$).
\paragraph{Elitism.}
To preserve high-performing solutions, a fixed number of elite genomes per non-stagnated species, referred to as the \emph{elite size}, are copied unchanged into the genome pool for the next generation (see \figref{fig:algorithm}(a)).

For stagnated species, instead of a fixed elite size, only a fraction of the top-performing individuals is preserved. Specifically, the number of survivors is given by
\[
\max\!\left(1,\; \left\lfloor \rho_{\text{elite}} \cdot |\mathcal{C}| \right\rfloor \right),
\]
where $\rho_{\text{elite}}$ denotes the \emph{elite survival fraction} and $|\mathcal{C}|$ is the species size. This ensures that at least one representative of each species is retained, preserving high-quality circuits while gradually suppressing stagnated species.

\paragraph{Offspring Allocation.}
Offspring allocation determines how many of the remaining $N-K$ individuals each non-stagnated species contributes to the next generation, where $K$ is the total number of elites. Since fitness values may be negative, we use a linear rank-based scheme rather than raw fitness. Reproducing species are ranked by their best fitness; if there are $R$ such species, 
the highest-ranked species receives weight $R$, down to $1$ for the lowest, and offspring counts are assigned proportionally to these weights (see the highlighted offspring counts for the non-stagnated species in \figref{fig:algorithm}(b)). Unlike NEAT, which allocates offspring counts based on average fitness of species, SPECS decides them based on best fitness. This is motivated by the highly combinatorial and ``spiky'' nature of the analog circuit search space, where a high average fitness often indicates many genetically similar solutions with limited diversity. In contrast, rewarding species based on their best individual encourages variation and better aligns with the task of finding the best possible solution. Empirically, we observe that many high-performing circuits are discovered by selecting the previously best circuit as parent, further supporting this design choice.

\paragraph{Parent Selection and Reproduction.}

Within each reproducing species, offspring are generated as follows. For each offspring, we first decide whether to perform crossover according to the \emph{crossover probability}. One parent is selected if crossover is skipped, or two parents if crossover is performed, using linear rank-based weighting. Only the top-performing fraction of individuals within the species, defined by the \emph{parent selection fraction} ($\rho_{\text{parent}}$), are eligible as parents. The selected parents produce an intermediate offspring via crossover; if crossover is skipped, the single parent is treated as the intermediate offspring. Finally, the intermediate offspring may undergo one structural mutation and/or one sizing mutation with predefined probabilities to produce the final offspring. This process repeats until the required $N - K$ offspring are generated.

\paragraph{Parallel-resistor simplification.}
As a lightweight topology-cleanup step applied to each new offspring, multiple resistors connecting the same pair of nets are replaced by a single equivalent resistor. The resistor with the smallest component ID is retained with its resistance updated to the equivalent parallel resistance, while the remaining resistors and their associated genes are removed. This reduces topological bloat without changing circuit behavior.

\paragraph{Speciation of the New Generation.}
Once the genome pool for the next generation is complete, each genome is assigned to a species using the same distance-based procedure as in initialization (see \figref{fig:algorithm}(c)). If the minimum distance to any species representative is below the compatibility threshold $\delta_t$, the genome joins the closest species; otherwise, it forms a new species. This is illustrated in \figref{fig:algorithm}(c)--(d), where a genome whose distance to all existing representatives exceeds $\delta_t$ forms a new species with ID 4. The representative of each species is then chosen as the medoid, i.e., the genome that minimizes the total distance to all other members.

\paragraph{Fitness Evaluation and Updates.}
Each genome is evaluated using Ngspice simulations to compute its task-specific fitness. Afterward, species best fitness values, stagnation counters, and representatives are updated (see \figref{fig:algorithm}(d)).

\paragraph{Adaptive Speciation Control.}
To maintain a target number of species $S$, the compatibility threshold $\delta_t$ is adjusted multiplicatively after each generation: it is multiplied by $1.05$ if the number of species exceeds the target, by $1/1.05 \approx 0.952$ if it is below the target, and left unchanged otherwise.

\paragraph{Termination.}

The evolutionary process continues until the predefined Ngspice simulation budget is exhausted. The circuit represented by the genome with the best fitness encountered during the entire run is returned as the final synthesized circuit.

\section{Experiments}
\label{sec:experiments}
\subsection{Task Definitions}
We evaluate SPECS on a set of computational analog circuit synthesis tasks widely used in the evolutionary circuit design literature~\cite{comp_1, comp_2, comp_3, comp_4, comp_5, acid_ge, acid_mge}, adopting exactly the same task definitions, simulation setup, and evaluation protocol as in ~\cite{acid_mge}.

The circuits implement four mathematical functions of the input voltage: squaring ($V_{\text{out}} = V_{\text{in}}^2$), square root ($V_{\text{out}} = \sqrt{V_{\text{in}}}$), cubing ($V_{\text{out}} = V_{\text{in}}^3$), and cube root ($V_{\text{out}} = \sqrt[3]{V_{\text{in}}}$). For squaring, cubing, and cube root, the input varies within $[-250\,\text{mV},\,250\,\text{mV}]$; for square root, within $[0\,\text{V},\,500\,\text{mV}]$. In each case, $M = 21$ uniformly spaced fitting points are evaluated.

Let $\tilde{V}_i$ and $V_i$ denote the ideal and simulated output voltages at fitting point $i$, and let $V_{\max} := \max_{1 \le i \le M} |\tilde{V}_i|$. A fitting point $i$ is a \emph{hit} if
\[
|V_i - \tilde{V}_i| \le 0.05 \, V_{\max},
\]
and a circuit is a \emph{success} if all $M = 21$ points are hits. The fitness used during evolution is based on the weighted absolute error
\[
E := \sum_{i=1}^{M} w_i \, |V_i - \tilde{V}_i|, \quad
w_i :=
\begin{cases}
1, & \text{if point } i \text{ is a hit,} \\
10, & \text{otherwise,}
\end{cases}
\]
where we define $fitness := 1 - E$. For reporting purposes, we also compute the mean absolute error
\[
\text{MAE} := \frac{1}{M} \sum_{i=1}^{M} |V_i - \tilde{V}_i|.
\]

All circuits are evaluated within a fixed test fixture with $\pm\SI{15}{\volt}$ supplies, a \SI{1}{\kilo\ohm} series input resistor, and a \SI{1}{\kilo\ohm} output load resistor to ground. The component library consists of default SPICE NPN and PNP BJT models and resistors with values ranging from $1\,\Omega$ to $10^9\,\Omega$, rounded to one decimal place. 

To encourage compact circuit designs and prevent uncontrolled topology growth, SPECS enforces an upper bound on the number of components per genome. Once this limit is reached, the \emph{Add Component} mutation is disabled for that genome. For all tasks, the maximum number of components is set to 50.

Circuits are evaluated using Ngspice transient simulations with a voltage ramp input of rise time $0.2\,\text{s}$, consistent with~\cite{acid_mge}.

\subsection{Parameter Configuration}
The three main runtime hyperparameters of SPECS are \emph{population size} $N$, \emph{target species number} $S$, and \emph{parent selection fraction} $\rho_{\text{parent}}$. To select these values, we performed a three-dimensional grid search over population sizes $\{50,100,200,400\}$, target species numbers $\{1,2,4,8\}$, and selection fractions $\{0.1,0.25,0.5,0.65\}$, yielding a total of $4\times4\times4=64$ configurations. Each configuration was evaluated on all four tasks using five independent runs with a budget of $10^{6}$ Ngspice simulations.

For every configuration, we record the median of the minimum MAE achieved across the five runs, yielding one scalar per task. Because the absolute MAE scales differ across tasks, the four per-task scores cannot be averaged directly. We therefore standardize the 64 median-MAE values within each task by $z$-normalization, and average the four normalized scores to obtain a single comparable score $\bar{z}$ for each configuration. The configuration with the lowest $\bar{z}$ is selected. \tabref{tab:grid_search} reports the resulting $\bar{z}$ scores for the full grid. The raw per-task median minimum MAE values used to compute the $\bar{z}$ scores are provided in the supplementary repository\footref{fn_repo}.

\begin{table}[t]
\centering
\footnotesize
\setlength{\tabcolsep}{8pt}
\renewcommand{\arraystretch}{1.15}
\caption{Normalized grid-search $\bar{z}$ scores (lower is better), averaged
over the four tasks, for each configuration of parent selection fraction ($\rho_{parent}$),
population size ($N$), and target species number ($S$). The best
configuration is shown in bold.}
\label{tab:grid_search}
\vspace{-0.3cm}
\begin{tabular}{cc rrrr}
\toprule
$\rho_{parent}$ & $N$ &
\multicolumn{1}{c}{$S{=}1$} &
\multicolumn{1}{c}{$S{=}2$} &
\multicolumn{1}{c}{$S{=}4$} &
\multicolumn{1}{c}{$S{=}8$} \\
\midrule
\multirow{4}{*}{0.10}
 & 50  & \cellcolor{cellred!17} 0.37 & \cellcolor{cellred!58} 1.26 & \cellcolor{cellred!22} 0.49 & \cellcolor{cellgreen!22} $-$0.23 \\
 & 100  & \cellcolor{cellred!58} 1.26 & \cellcolor{cellred!32} 0.70 & \cellcolor{cellgreen!30} $-$0.31 & \cellcolor{cellred!8} 0.17 \\
 & 200  & \cellcolor{cellred!33} 0.72 & \cellcolor{cellred!25} 0.54 & \cellcolor{cellgreen!30} $-$0.32 & \cellcolor{cellgreen!72} $-$0.76 \\
 & 400  & \cellcolor{cellred!1} 0.03 & \cellcolor{cellred!50} 1.08 & \cellcolor{cellgreen!48} $-$0.50 & \cellcolor{cellgreen!74} $-$0.78 \\
\midrule
\multirow{4}{*}{0.25}
 & 50  & \cellcolor{cellgreen!2} $-$0.02 & \cellcolor{cellgreen!30} $-$0.31 & \cellcolor{cellred!13} 0.28 & \cellcolor{cellred!37} 0.81 \\
 & 100  & \cellcolor{cellgreen!16} $-$0.17 & \cellcolor{cellgreen!16} $-$0.17 & \cellcolor{cellgreen!7} $-$0.07 & \cellcolor{cellgreen!62} $-$0.65 \\
 & 200  & \cellcolor{cellgreen!29} $-$0.30 & \cellcolor{cellgreen!53} $-$0.56 & \cellcolor{cellgreen!21} $-$0.22 & \cellcolor{cellgreen!46} $-$0.48 \\
 & 400  & \cellcolor{cellgreen!35} $-$0.37 & \cellcolor{cellgreen!28} $-$0.29 & \cellcolor{cellgreen!100} \textbf{$-$1.05} & \cellcolor{cellgreen!52} $-$0.55 \\
\midrule
\multirow{4}{*}{0.50}
 & 50  & \cellcolor{cellred!3} 0.06 & \cellcolor{cellgreen!34} $-$0.36 & \cellcolor{cellgreen!10} $-$0.10 & \cellcolor{cellgreen!8} $-$0.08 \\
 & 100  & \cellcolor{cellred!11} 0.23 & \cellcolor{cellgreen!17} $-$0.18 & \cellcolor{cellgreen!39} $-$0.41 & \cellcolor{cellgreen!11} $-$0.12 \\
 & 200  & \cellcolor{cellred!6} 0.12 & \cellcolor{cellgreen!19} $-$0.20 & \cellcolor{cellgreen!58} $-$0.61 & \cellcolor{cellgreen!28} $-$0.29 \\
 & 400  & \cellcolor{cellred!33} 0.71 & \cellcolor{cellgreen!13} $-$0.14 & \cellcolor{cellgreen!27} $-$0.28 & \cellcolor{cellgreen!32} $-$0.34 \\
\midrule
\multirow{4}{*}{0.65}
 & 50  & \cellcolor{cellgreen!3} $-$0.03 & \cellcolor{cellgreen!30} $-$0.32 & \cellcolor{cellred!7} 0.16 & \cellcolor{cellred!14} 0.30 \\
 & 100  & \cellcolor{cellred!11} 0.25 & \cellcolor{cellgreen!6} $-$0.06 & \cellcolor{cellred!17} 0.36 & \cellcolor{cellgreen!15} $-$0.16 \\
 & 200  & \cellcolor{cellred!100} 2.18 & \cellcolor{cellgreen!25} $-$0.26 & \cellcolor{cellgreen!20} $-$0.21 & \cellcolor{cellred!5} 0.10 \\
 & 400  & \cellcolor{cellred!52} 1.14 & \cellcolor{cellgreen!44} $-$0.46 & \cellcolor{cellgreen!18} $-$0.19 & \cellcolor{cellgreen!42} $-$0.44 \\
\bottomrule
\end{tabular}
\vspace{-0.3cm}
\end{table}
 
The target species number $S$ deserves special attention. Setting $S=1$ effectively removes the role of speciation, since the population is driven toward a single species, and therefore serves as a useful ablation. \tabref{tab:grid_search} shows that $S=1$ yields the highest (worst) $\bar{z}$ scores overall, with averages of $0.39$, $0.02$, $-0.19$, and $-0.22$ for $S=1$, $2$, $4$, and $8$, respectively, reflecting a consistent improvement with increasing $S$. This result confirms that speciation is a key component of the proposed method, supporting the intuition that it preserves diversity and avoids premature loss of promising innovations.

The configuration with $N=400$, $S=4$, and $\rho_{\text{parent}}=0.25$ achieved the lowest $\bar{z}=-1.05$ and is therefore used in all experiments reported in this paper. All remaining hyperparameters were fixed to values based on intuition and preliminary experimentation, and were not further optimized. \tabref{tab:parameters} summarizes the final parameter settings.

\begin{table}[t]
\centering
\footnotesize
\setlength{\tabcolsep}{5pt}
\renewcommand{\arraystretch}{1.25}
\caption{Parameter configuration used in all experiments.}
\label{tab:parameters}
\vspace{-0.3cm}
\begin{tabular}{lr|lr}
\toprule
\multicolumn{4}{c}{\textit{Population \& Speciation}} \\[4pt]
Population size ($N$)            & 400  & Compatibility threshold ($\delta_t$)        & 0.75 \\
Target species number ($S$)      & 4    & Stagnation threshold                   & 10   \\
Parent selection frac. ($\rho_{\text{parent}}$) & 0.25 & Elite size              & 1    \\
Elite survival frac. ($\rho_{\text{elite}}$)    & 0.1  &                          &      \\
\midrule
\multicolumn{4}{c}{\textit{Reproduction \& Global}} \\[4pt]
Crossover prob.            & 0.5  & Structural mutation prob. & 0.8  \\
Sizing mutation prob.      & 0.9  & Max number of comp.            & 50   \\
\midrule
\multicolumn{4}{c}{\textit{Structural Mutation Types}} \\[4pt]
Add component prob.        & 0.25 & Split net prob.           & 0.25 \\
Delete component prob.     & 0.25 & Rewire prob.              & 0.25 \\
\midrule
\multicolumn{4}{c}{\textit{Sizing Mutation Types \& Add Component Modes}} \\[4pt]
Resample param.\ prob.     & 0.5  & Zero-detachment prob.     & 1/3  \\
Perturb param.\ prob.      & 0.5  & One-detachment prob.      & 1/3  \\
                           &      & Two-detachment prob.      & 1/3  \\
\midrule
\multicolumn{4}{c}{\textit{Distance Coefficients}} \\[4pt]
Excess connections ($c_1$) & 0.25 & Param.\ mismatch ($c_3$) & 0.25 \\
Disjoint connections ($c_2$) & 0.25 & Comp.\ mismatch ($c_4$) & 0.25 \\
\bottomrule
\end{tabular}
\vspace{-0.3cm}
\end{table}

\subsection{Benchmark Methods}

\captionsetup[table*]{justification=centering}
\begin{table*}[!t]
\centering
\footnotesize
\setlength{\tabcolsep}{5pt}
\renewcommand{\arraystretch}{1.1}
\caption{SPECS vs.\ GraCo-ES vs.\ SPICEMixer vs. SPICEMixer++ for the computational circuit tasks.\\
All results are reported over 5 runs with random seeds, each with a budget of 
$\mathbf{10^{6}}$ Ngspice simulations.}
\label{tab:results_1}
\vspace{-0.3cm}
\begin{tabular}{llccccccc}
\toprule
Circuit & Algorithm &
SR (\%) &
Mean Hits (\%) $\pm$ SD &
MAE $\pm$ SD (mV) &
Mean BF $\pm$ SD &
Max BF &
Min MAE (mV) &
NCBC \\
\midrule

\multirow{4}{*}{Squaring}
 & GraCo-ES & 0 & 23.8 $\pm$ 9.5 & 16.86 $\pm$ 1.22 & -2.44 $\pm$ 0.26 & -1.928 & 14.41 & \textbf{6} \\
 & SPICEMixer & 20 & 52.4 $\pm$ 37.0 & 9.87 $\pm$ 6.82 & -0.90 $\pm$ 1.51 & 0.974 & 1.24 & 27 \\
 & SPICEMixer++ & 20 & 83.8 $\pm$ 14.3 & 4.83 $\pm$ 4.26 & 0.19 $\pm$ 0.88 & \textbf{0.992} & 0.38 & 50 \\
& SPECS & \textbf{80} & $\mathbf{98.1 \pm 3.8}$ & $\mathbf{0.76 \pm 0.51}$ & $\mathbf{0.93 \pm 0.07}$ & 0.991 &\textbf{0.17} & 47 \\
\midrule

\multirow{4}{*}{Square root}
 & GraCo-ES & 0 & 18.1 $\pm$ 5.6 & 99.44 $\pm$ 7.77 & -19.26 $\pm$ 1.86 & -16.086 & 86.62 & \textbf{3} \\
 & SPICEMixer & 20 & 42.9 $\pm$ 29.8 & 79.96 $\pm$ 46.57 & -14.90 $\pm$ 9.96 & 0.904 & 4.56 & 22 \\
 & SPICEMixer++ & \textbf{60} & 83.8 $\pm$ 24.0 & 25.62 $\pm$ 24.12 & -2.66 $\pm$ 4.88 & 0.906 & 0.99 & 241 \\
  & SPECS & \textbf{60} & $\mathbf{98.1 \pm 2.3}$ & $\mathbf{4.44 \pm 4.21}$ & $\mathbf{0.56 \pm 0.49}$ & \textbf{0.989} & \textbf{0.40} & 47 \\
\midrule

\multirow{4}{*}{Cubing}
 & GraCo-ES & 0 & 21.9 $\pm$ 2.3 & 2.25 $\pm$ 0.01 & 0.55 $\pm$ 0.00 & 0.547 & 2.24 & \textbf{6} \\
 & SPICEMixer & 20 & 95.2 $\pm$ 3.0 & 0.43 $\pm$ 0.15 & 0.97 $\pm$ 0.02 & 0.996 & 0.19 & 36 \\
 & SPICEMixer++ & 40 & 90.5 $\pm$ 12.4 & 0.40 $\pm$ 0.35 & 0.95 $\pm$ 0.06 & 0.998 & 0.10 & 50 \\
  & SPECS & \textbf{100} & $\mathbf{100.0 \pm 0.0}$ & $\mathbf{0.06 \pm 0.03}$ & $\mathbf{1.00 \pm 0.00}$ & \textbf{0.999} & \textbf{0.03} & 50 \\
\midrule

\multirow{4}{*}{Cube root}
 & GraCo-ES & 0 & 4.8 $\pm$ 0.0 & 407.00 $\pm$ 0.05 & -84.47 $\pm$ 0.01 & -84.464 & 406.97 & \textbf{2} \\
 & SPICEMixer & 0 & 78.1 $\pm$ 10.7 & 23.65 $\pm$ 6.57 & -1.99 $\pm$ 1.39 & 0.196 & 13.31 & 32 \\
 & SPICEMixer++ & 40 & 70.5 $\pm$ 37.4 & 125.21 $\pm$ 158.40 & -24.24 $\pm$ 33.74 & 0.788 & 5.36 & 288 \\
  & SPECS & \textbf{100} & $\mathbf{100.0 \pm 0.0}$ & $\mathbf{3.88 \pm 1.51}$ & $\mathbf{0.24 \pm 1.08}$ & \textbf{0.949} & \textbf{2.14} & 48 \\
\bottomrule
\end{tabular}
\end{table*}

\captionsetup[table*]{justification=centering}
\begin{table*}[!t]
\centering
\footnotesize
\setlength{\tabcolsep}{5pt}
\renewcommand{\arraystretch}{1.1}
\caption{SPECS vs.\ ACID-GE vs.\ ACID-MGE for the computational circuit tasks.\\
All results are reported over 50 runs with random seeds, each with a budget of 
$\mathbf{3 \times 10^{6}}$ Ngspice simulations.}
\label{tab:results_2}
\vspace{-0.3cm}
\begin{tabular}{llccccccc}
\toprule
Circuit & Algorithm &
SR (\%) &
Mean Hits (\%) $\pm$ SD &
MAE $\pm$ SD (mV) &
Mean BF $\pm$ SD &
Max BF &
Min MAE (mV) &
NCBC \\
\midrule

\multirow{3}{*}{Squaring}
 & ACID-GE & 52 & 81.0 $\pm$ 28.5 & 0.93 $\pm$ 0.41 & 0.27 $\pm$ 1.17 & 0.991 & -- & \textbf{28} \\
 & ACID-MGE & 84 & 97.7 $\pm$ 6.5 & 0.53 $\pm$ 0.24 & 0.94 $\pm$ 0.14 & \textbf{0.998} & \textbf{0.08} & 42 \\
  & SPECS & \textbf{100} & $\mathbf{100.0 \pm 0.0}$ & $\mathbf{0.34 \pm 0.17}$ & $\mathbf{0.99 \pm 0.00}$ & \textbf{0.998} & 0.09 & 45 \\
\midrule

\multirow{3}{*}{Square root}
 & ACID-GE & 44 & 74.0 $\pm$ 32.3 & 5.55 $\pm$ 4.32 & -5.25 $\pm$ 7.61 & 0.972 & -- & \textbf{35} \\
 & ACID-MGE & 66 & 89.0 $\pm$ 24.2 & 4.01 $\pm$ 2.47 & -1.71 $\pm$ 5.78 & \textbf{0.997} & 0.23 & 44 \\
  & SPECS & \textbf{90} & $\mathbf{99.2 \pm 2.6}$ & $\mathbf{3.68 \pm 7.62}$ & $\mathbf{0.73 \pm 1.22}$ & 0.996 & \textbf{0.18} & 49 \\
\midrule

\multirow{3}{*}{Cubing}
 & ACID-GE & 76 & 92.2 $\pm$ 19.3 & 0.18 $\pm$ 0.07 & 0.95 $\pm$ 0.12 & 0.998 & -- & \textbf{40} \\
 & ACID-MGE & 84 & 95.7 $\pm$ 13.7 & 0.10 $\pm$ 0.04 & 0.97 $\pm$ 0.08 & \textbf{0.999} & 0.05 & 47 \\
  & SPECS & \textbf{92} & $\mathbf{99.0 \pm 5.4}$ & $\mathbf{0.06 \pm 0.03}$ & $\mathbf{1.00 \pm 0.00}$ & \textbf{0.999} & \textbf{0.02} & 48 \\
\midrule

\multirow{3}{*}{Cube root}
 & ACID-GE & 2 & 11.5 $\pm$ 20.2 & 10.81 $\pm$ 0.00 & -75.95 $\pm$ 23.74 & 0.773 & -- & \textbf{41} \\
 & ACID-MGE & 22 & 70.2 $\pm$ 29.4 & 7.13 $\pm$ 3.70 & -12.87 $\pm$ 25.02 & 0.964 & 2.04 & 57 \\
  & SPECS & \textbf{100} & $\mathbf{100.0 \pm 0.0}$ & $\mathbf{5.56 \pm 4.80}$ & $\mathbf{0.77 \pm 0.49}$ & \textbf{0.970} & \textbf{1.41} & 48 \\
\bottomrule
\end{tabular}
\vspace{-0.1cm}
\end{table*}

We compare SPECS against four benchmark methods from the recent literature that jointly address circuit topology synthesis and component sizing. All methods use the same test fixture, simulation setup, and evaluation metrics, ensuring that observed differences are attributable to the synthesis methods themselves.

GraCo~\cite{graco_es} formulates circuit synthesis as an autoregressive graph-construction problem. Circuits are represented as graphs and generated step by step before being translated into SPICE netlists for simulation. The method supports two search strategies: REINFORCE with a leave-one-out baseline (RLOO) and evolution strategies (ES). We use the ES variant, which the original paper reports to yield better results than RLOO on the considered tasks.

SPICEMixer~\cite{spicemixer} addresses the same joint topology-and-sizing problem but through a substantially different representation. While GraCo generates circuits through an intermediate graph representation, SPICEMixer operates directly on normalized SPICE netlists, applying crossover, mutation, and pruning at the level of netlist lines without the need for a separate handcrafted chromosome encoding. We also benchmark against a further enhanced variant of SPICEMixer, which increases the elite set size from 30 to 100 and uses a richer set of genetic operators, providing broader diversity preservation and finer-grained exploration of the netlist search space. We refer to this variant as SPICEMixer++.

ACID-GE~\cite{acid_ge} is a grammar-based evolutionary approach that uses grammatical evolution to decode variable-length chromosomes into analog circuit netlists, coupling topology generation and sizing through a grammar-guided genotype-to-phenotype mapping within a single evolutionary process.

ACID-MGE~\cite{acid_mge} extends ACID-GE by introducing modularity and homology through multi-grammatical evolution. The chromosome is partitioned into separate parts associated with different subproblems, each decoded by its own grammar, reducing the disruptive effect of crossover and improving the reuse of beneficial substructures. According to reported results, ACID-MGE improves over ACID-GE in success rate and MAE on the computational circuit tasks, though often at the cost of increased circuit complexity.

Overall, these benchmarks represent distinct paradigms for joint topology-and-sizing optimization: graph-based generation in GraCo-ES, direct netlist-level evolution in SPICEMixer and SPICEMixer++, and grammar-based evolution in ACID-GE and ACID-MGE.

\subsection{Results}

\tabref{tab:results_1} and \tabref{tab:results_2} summarize the results. Following~\cite{acid_mge}, SR denotes the success rate, NCBC the number of components in the best circuit, and BF the best fitness.

\begin{figure*}[!t]
    \resizebox{\linewidth}{!}{%
    \includegraphics[trim=0 20 0 0]{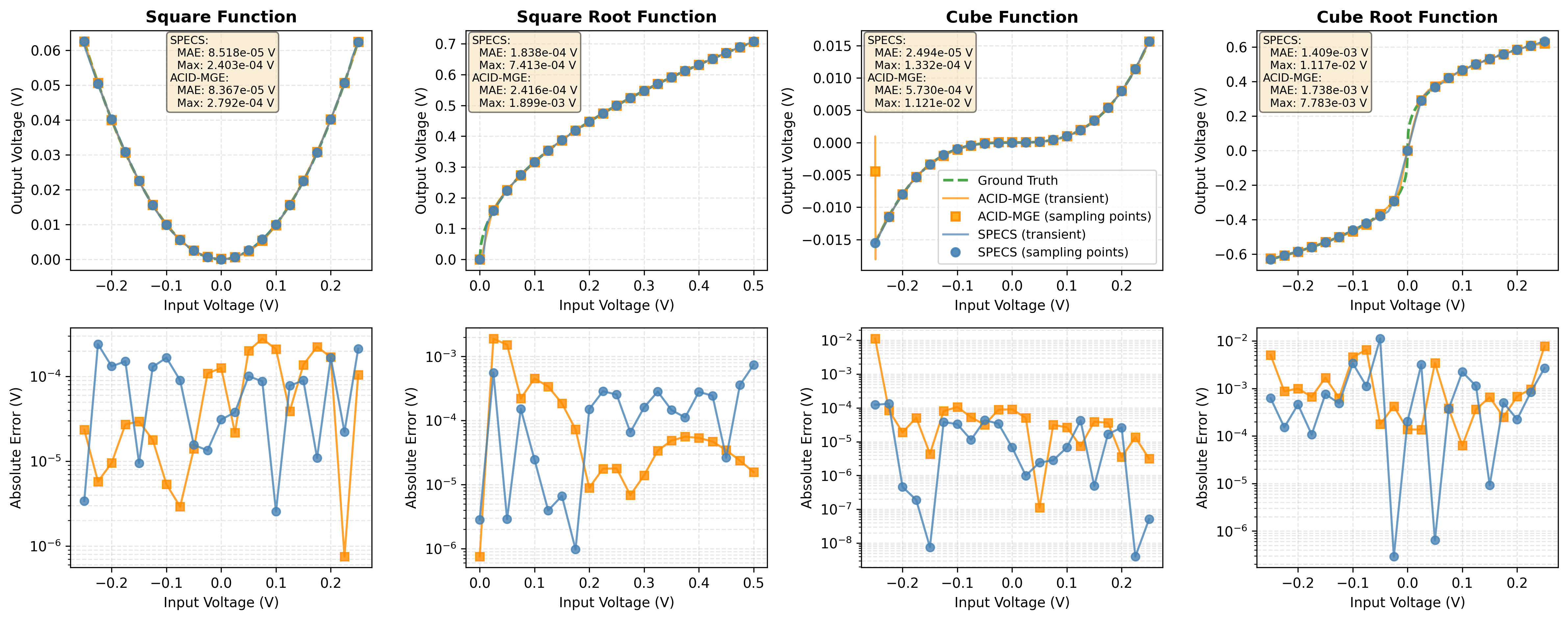}}
    \caption{Transient responses of the best circuits found by SPECS and ACID-MGE on the computational circuit synthesis tasks.}
    \label{fig:math_functions_waveforms}
    \vspace{-0.4cm}
\end{figure*}

To provide an initial reference for performance, \tabref{tab:results_1} compares SPECS against GraCo-ES, SPICEMixer, and SPICEMixer++ using 5 independent runs with $10^6$ Ngspice simulations each, the same budget used for hyperparameter search. GraCo-ES fails to synthesize any successful circuit, consistently collapsing to solutions with very few components (NCBC of 2 to 6) and highly negative mean BF values, indicating premature convergence to structurally degenerate circuits. SPICEMixer and SPICEMixer++ perform substantially better but still fall well short of SPECS across all aggregate metrics on all tasks. In terms of the best circuits found, SPECS achieves higher Max BF and lower Min MAE on every task, with the only exceptions being squaring and cubing, where it is effectively tied with SPICEMixer++ in Max BF.

\tabref{tab:results_2} compares SPECS against ACID-GE and ACID-MGE following the evaluation protocol of~\cite{acid_mge}, where each method is evaluated over 50 independent runs with a budget of $3\times10^6$ Ngspice simulations, enabling direct comparison with published results. SPECS improves over ACID-GE on all metrics except NCBC, across all tasks. The more demanding comparison is against ACID-MGE, the strongest benchmark method overall, which SPECS still consistently outperforms in all aggregate metrics across all tasks. The largest gap appears on cube root, where SPECS reaches 100\% SR against ACID-MGE's 22\%. On squaring, square root, and cubing, the two methods are effectively tied in Max BF, indicating that ACID-MGE can occasionally find circuits of comparable peak quality; nevertheless, SPECS attains lower Min MAE on three tasks of the four. Notably, these gains are not explained by larger circuits: the average NCBC over all four tasks is 47.5 for both methods.

\figref{fig:math_functions_waveforms} provides the transient responses of the best circuits found by SPECS and ACID-MGE on all four tasks. Both methods closely approximate the target functions across the full input range. On squaring and square root tasks, the two circuits perform comparably, consistent with the near-tied results in \tabref{tab:results_2}. On cubing and cube root tasks, however, SPECS achieves a more accurate fit on most fitting points.

The SPICE netlists of the best circuits found by SPECS reported
in \tabref{tab:results_2}, together with their schematics and
transient simulation testbenches, are provided in the supplementary repository\footref{fn_repo}.

The experimental results demonstrate that SPECS achieves the best single-run performance on most tasks and effectively ties the strongest benchmark methods on the rest. Beyond best single-run performance, its most important advantage is reliability: it discovers successful circuits much more consistently across repeated runs than any considered benchmark, as reflected by substantially higher SR and mean hit rates and lower mean MAE. We attribute this to three main factors. First, starting from minimal structures and increasing complexity only through evolutionary improvement ensures that genes present in the population have earned their place, avoiding wasted effort on structurally unjustified combinations. Second, wiring-constraint-aware mutation operators guarantee that all offspring represent electrically valid topologies, avoiding wasted simulations on infeasible candidates. Third, speciation allows novel solutions to compete within their own niches rather than being immediately exposed to more mature topologies, reducing the risk that promising innovations are discarded before they can be refined. This maintains diversity, mitigates premature convergence, and makes the search less likely to become trapped in suboptimal regions of the design space.

\section{Conclusion}
\label{sec:conclusion}

In this paper, we introduced SPECS, a novel evolutionary algorithm for analog circuit synthesis that jointly searches over circuit topology and component parameters. The method combines a circuit-native genome representation with topology-aware crossover and mutation operators, wiring constraints, and a speciation-based search mechanism that preserves structural innovation while allowing complexity to grow only when beneficial. A key element is the historical tracking of structural changes through innovation IDs, inspired by NEAT, enabling a meaningful distance metric between genomes and supporting effective speciation throughout search.

We evaluated SPECS on four computational analog circuit synthesis tasks and demonstrated consistent improvements over benchmark methods from graph-based, netlist-level, and grammar-based synthesis paradigms in both best single-run and aggregate performance. The key strength of SPECS lies in three main design principles: minimal initialization that grows complexity only through evolutionary improvement, a domain-specific set of genetic operators that enforce wiring constraints and keep the search within the space of electrically valid circuits, and speciation that protects structural diversity and prevents premature convergence. Together, these make SPECS a reliable and efficient synthesizer that performs strongly not just on the best run, but consistently across independent runs. The fact that the same parameter configuration proved effective across all tasks without extensive tuning further suggests that these principles generalize well beyond the specific problems considered here.

The current evaluation focused on BJTs and resistors, but SPECS is fully compatible with arbitrary component types and fitness functions, making it a general framework for automated analog design. Future work could extend the evaluation to richer component libraries including MOS transistors, capacitors, inductors, and diodes, as well as more application-oriented tasks such as amplifiers, filters, and oscillators. This would put the generalizability of the approach to a broader test and further establish its potential as a widely applicable tool for automated analog circuit synthesis.

\ifthenelse{\boolean{anonymous}}{}{%
\section*{Acknowledgements}
The authors would like to thank Phuoc Pham (TU Munich) for his valuable assistance in generating the circuit schematics.
}

\clearpage

\bibliographystyle{IEEEtran}
\bibliography{references}

\end{document}